\pdfoutput=1

\documentclass[11pt]{article}

\usepackage[preprint]{acl}

\usepackage{times}
\usepackage{latexsym}

\usepackage[T1]{fontenc}

\usepackage[utf8]{inputenc}

\usepackage{microtype}

\usepackage{inconsolata}
\usepackage{adjustbox}
\usepackage{booktabs}
\usepackage{hyperref}
\usepackage{multirow}
\usepackage{listings}
\usepackage[ruled, vlined, linesnumbered]{algorithm2e}

\usepackage{amsmath}
\usepackage{times}
\usepackage{framed}
\usepackage{ragged2e}
\usepackage{bm}
\usepackage{soul}
\usepackage{latexsym}
\usepackage{subfigure}
\usepackage{amsthm}
\usepackage{graphicx}
\usepackage{float}
\usepackage{longtable}
\usepackage{booktabs} 
\usepackage{multirow}
\usepackage{multicol}
\usepackage{amssymb}
\usepackage{dashbox}%
\usepackage{multirow}
\usepackage{textcomp}
\usepackage{tcolorbox}
\usepackage{makecell}
\usepackage{bbding}
\usepackage[ruled, vlined, linesnumbered]{algorithm2e}
\usepackage{mathtools}
\usepackage{adjustbox}
\usepackage{color, soul}
\usepackage{pifont}
\usepackage{listings}
\usepackage{cleveref}
\crefname{section}{§}{§§}
\Crefname{section}{§}{§§}
\crefname{figure}{Figure}{Figure}
\Crefname{figure}{Figure}{Figure}
\crefname{table}{Table}{Table}
\Crefname{table}{Table}{Table}
\definecolor{my_red}{RGB}{255,99,71}
\definecolor{my_green}{RGB}{50,205,50}
\definecolor{my_blue}{RGB}{65,105,225}
\definecolor{forestgreen}{HTML}{228B22}

\definecolor{codegreen}{rgb}{0,0.6,0}
\definecolor{codegray}{rgb}{0.5,0.5,0.5}
\definecolor{codepurple}{rgb}{0.58,0,0.82}
\definecolor{backcolour}{rgb}{0.95,0.95,0.92}

\lstdefinestyle{mystyle}{
    backgroundcolor=\color{backcolour},   
    commentstyle=\color{codegreen},
    stringstyle=\color{codepurple},
    basicstyle=\ttfamily\scriptsize,
    breakatwhitespace=true,         
    breaklines=true,                 
    captionpos=b,                    
    keepspaces=true,                 
    numbers=none,                    
    numbersep=5pt,                  
    showspaces=false,                
    showstringspaces=false,
    showtabs=false,                  
    tabsize=2,
    columns=flexible,
    escapeinside={(*}{*)},
}

\title{Beyond Scaling: Predicting Patent Approval with Domain-specific Fine-grained Claim Dependency Graph}

\author{Xiaochen Kev Gao$^{1*}$, Feng Yao$^{1*}$, Kewen Zhao$^{2}$, Beilei He$^{3}$, \\ \textbf{Animesh Kumar$^{1}$, Vish Krishnan$^{1}$, Jingbo Shang$^{1}$}\vspace{0.2em} \\
$^1$ University of California San Diego,\\
$^2$ Carnegie Mellon University,
$^3$ University of Pennsylvania\vspace{0.2em}\\
{\tt \{xig034,fengyao,ank028,vkrishnan,jshang\}@ucsd.edu}\\
{\tt \{kewenz\}@cs.cmu.edu,  \{beilei\}@seas.upenn.edu}
}

\begin{document}
\maketitle
\begingroup\def\thefootnote{$*$}\footnotetext{The first two authors contributed equally to this paper. The listing order is fully decided by dice rolling.}\endgroup

\begin{abstract}
Model scaling is becoming the default choice for many language tasks due to the success of large language models (LLMs). However, it can fall short in specific scenarios where simple customized methods excel.
In this paper, we delve into the patent approval prediction task and unveil that simple domain-specific graph methods outperform enlarging the model, using the intrinsic dependencies within the patent data. 
Specifically, we first extend the embedding-based state-of-the-art (SOTA) by scaling up its backbone model with various sizes of open-source LLMs, then explore prompt-based methods to harness proprietary LLMs' potential, but find the best results close to random guessing, underlining the ineffectiveness of model scaling-up.
Hence, we propose a novel Fine-grained cLAim depeNdency (FLAN) Graph through meticulous patent data analyses, capturing the inherent dependencies across segments of the patent text. As it is model-agnostic, we apply cost-effective graph models to our FLAN Graph to obtain representations for approval prediction.
Extensive experiments and detailed analyses prove that incorporating FLAN Graph via various graph models consistently outperforms all LLM baselines significantly.
We hope that our observations and analyses in this paper can bring more attention to this challenging task and prompt further research into the limitations of LLMs. Our source code and dataset can be obtained from \url{https://github.com/ShangDataLab/FLAN-Graph}.
\end{abstract}

\section{Introduction}

\begin{figure}[t]
    \centering
    \includegraphics[width=0.95\linewidth]{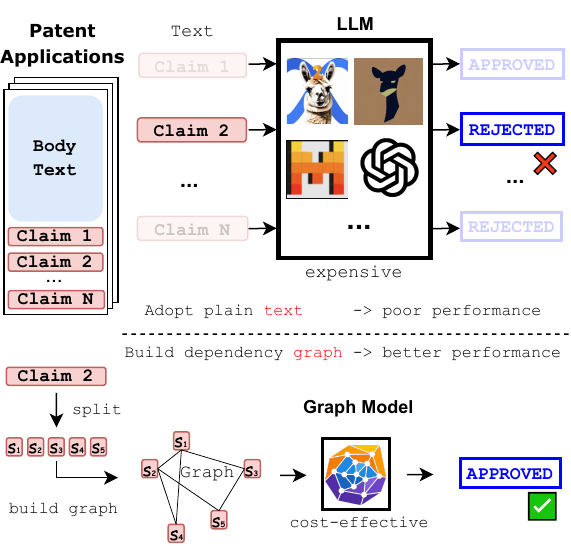}
    \caption{An illustration for the patent approval prediction task approached by LLMs and graph models, where each node of the graph is an informative segment decomposed from the original claim text.}
    \label{fig:intro}
\end{figure}

Scaling up language models has demonstrated predictable improvement and unprecedented abilities in many language tasks~\cite{chung2022scaling, wei2022emergent, zhao2023survey}. However, emerging evidence shows that simply scaling up backbone models to large language models (LLMs) may not guarantee success~\cite{peng2023does, hou2023large, wang2023doris}. In addition, scaling up models imposes demanding computational costs that prevent it from being widely adopted for real-world applications. Such limitations necessitate cost-effective methods beyond scaling, especially for domain-specific tasks that have distinct traits.

In this paper, we look into the task of patent approval prediction, a challenging yet straightforward classification task that scaling struggles to address, and explore customized cost-effective solutions. As shown in Figure~\ref{fig:intro}, the objective is to determine if each claim in a patent application will be approved or rejected by the U.S. government patent office (USPTO). It is vital for intellectual property (IP) protection, taking up to 40\% of the U.S. GDP and over 30\% of employment. Due to the demanding requirements for knowledge in both technology and law, patent examination is conducted manually, leading to potential inconsistent outcomes across patent examiners~\citep{inconsistency, ipeconomy}. Such inconsistency underscores the necessity for objective and automated computational support.

The state-of-the-art (SOTA) of this task is based on BERT embedding~\citep{devlin-etal-2019-bert} augmented with handcrafted features~\citep{gao2022towards}. An intuitive idea is to replace its backbone model with modern LLMs. To this end, we employ LLaMA2~\citep{touvron2023llama}, Mistral~\citep{jiang2023mistral}, Vicuna~\citep{chiang2023vicuna} in various sizes (7\&13\&70B) and apply both LoRA~\citep{hu2021lora} and full fine-tuning. Surprisingly, they do not live up to expectations, performing on par or worse than BERT. To exploit LLMs' emergent abilities, we utilize prompt-based methods tailored to these open-source LLMs, as well as closed-source GPT-3.5~\citep{openai2022chatgpt} and GPT-4~\citep{openai2023gpt4}, but the results are still unsatisfying.

\begin{figure}[t]
    \centering
    \includegraphics[width=0.9\linewidth]{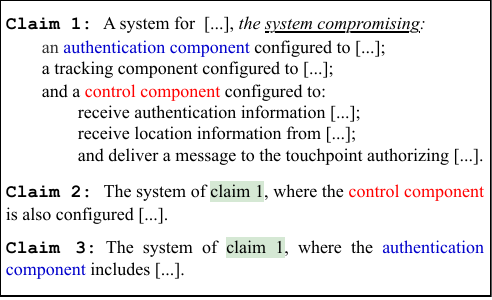}
    \caption{A brief example of the typical patent claim writing style and hierarchical dependencies within claims from a real-world patent application.}
    \label{fig:claim_short}
\end{figure}

The shattered hope in LLMs motivates us to dive into patent data analyses, which leads us to the standardized writing of claims and the dependencies nature among them. As depicted in Figure~\ref{fig:claim_short}, \textit{Claim 1} compromises multiple sub-components, which are then referenced in subsequent claims. Such intricate inner-claim (between sub-components in \textit{Claim 1}) and inter-claim dependencies (between \textit{Claims 1\&2} as well as \textit{Claims 1\&3}) have critical implications for the patent approval prediction task, as the patent examination is conducted on each claim and the rejection of one claim can result in the automatic rejection of its dependents.

Inspired by the observations and domain-specific knowledge acquired from painstakingly extensive data analyses, we propose Fine-grained cLAim depeNdency (FLAN) Graph for patent approval prediction, which represents each claim by a single graph that encapsulates both inner- and inter-claim dependencies. 
Specifically, as shown in Figure~\ref{fig:intro}, we first design a novel algorithm to automatically construct the FLAN Graph at scale, where each node is an informative segment of the claim text. Then, the model-agnostic FLAN Graph is fed into a generic graph model for prediction. Examples of the FLAN graphs and the corresponding claims are shown in Appendix~\ref{appx:full_example}.
In the experiments, we adopt a variety of cost-effective graph models such as GCN~\citep{chen2020simple}, GAT~\citep{velickovic2017graph}, and TreeLSTM~\citep{tai2015improved} to verify the effectiveness of our proposed FLAN Graph for patent approval prediction. All models with FLAN Graph applied outperform the previous SOTA, among which GraphSage~\citep{hamilton2017inductive} achieves the highest improvements of 7.4\% in AUC and 7.8\% in Macro-F1 scores, achieving absolute scores of 66.04 and 58.22, respectively.

To summarize, our contributions are two-fold: 
\textbf{(1)} We propose a novel algorithm to automatically construct the Fine-grained cLAim depeNdency (FLAN) Graph at scale that consistently improves the SOTA by a large margin.
\textbf{(2)} We conduct comprehensive experiments and analyses of modern LLMs on patent approval prediction, which identify the limitations of LLMs and provide valuable references for developing LLM-based solutions in the future.
The source code\footnote{\url{https://github.com/ShangDataLab/FLAN-Graph}} and dataset\footnote{\url{https://huggingface.co/datasets/ShangDataLab-UCSD/PatentAP}} are publicly released to facilitate future research.

\section{Problem Formulation}
In this section, we formally introduce the definition of the patent approval prediction task and analyze the dataset we construct for experiments.

\subsection{Task Definition}
\label{sec:task_definition}
As illustrated in Figure~\ref{fig:intro}, patent applications are initially submitted to the USPTO in the form of documents. The examination process, however, focuses on approving or rejecting each individual claim. Therefore, given a patent application $A_i = \{C^{(i)}_j\}^n_{j=1}$ containing $n$ claims, the task of patent approval prediction is to determine whether each claim $C_j^{(i)}$ will be approved or rejected by the USPTO indicated by a binary label $y_{j}^{(i)} \in \{0,1\}$. 

In practice, patent claims are reviewed according to the legal section 35 U.S. Code § 102, where the core criterion is novelty-based, bringing in distinct challenges below. (1) \textbf{Time-sensitive}. Unlike traditional text classification, novelty assessment depends on the application filing date, allowing opposite decisions for the same claim over time. (2) \textbf{Structure-dependent}. Many claims (e.g., \textit{Claims 2\&3} in Figure~\ref{fig:claim_short}) are dependent on others within the same application, and such structure can influence novelty evaluation. (3) \textbf{Knowledge-intensive}. Evaluating novelty requires up-to-date knowledge of both technologies and patent law. (4) \textbf{Outcome-inconsitent}. Novelty examination outcomes are subject to preferences across patent officers, which may introduce inconsistencies in patent data.

\subsection{Dataset Collection}
\label{sec:data_collect}
We collect data of real-world patent applications from \citet{gao2022towards} and filter out those outdated data before 2018. The data is initially merged and derived from the publicly available resources officially released on the USPTO websites\footnote{
\url{https://ped.uspto.gov/peds/\#/}
}. 

Considering the real-world scenario, we utilize historical data for training and more recent data for evaluation. Specifically, we sort the applications based on filing dates and then split them into training, validation, and test sets. As shown in Table~\ref{tab:data_stat}, the resulting dataset \textsc{PatentAP} is large-scale with about $1.5 M$ claims for training and $0.5 M$ for evaluation. It is also highly imbalanced, with most claims being approved, further adding to the difficulty. The claims are relatively short and 92\% have less than 128 tokens and the average length is 54. Each application has 17 claims on average.

\begin{table}[!t]
  \centering
  \small
  \begin{adjustbox}{max width=0.9\linewidth}
  {
      \begin{tabular}{c|rcc}
      \toprule
                & \textbf{\#Claim} &  \textbf{\#Application} & \textbf{Approval (\%)} \\
      \midrule
      \textbf{Train }    & $1,485,693$ & $87,883$& $81.36$ \\
      \textbf{Valid}     & $278,215$ & $16,955$& $83.41$ \\
      \textbf{Test }     & $185,477$ & $11,148$& $84.92$\\
      \bottomrule
      \end{tabular}
  }
  \end{adjustbox}
  \caption{Statistics of \textsc{PatentAP} dataset. ``Approval (\%)'' indicates the percentage of the approved claims.}
  \label{tab:data_stat}%
\end{table}

\section{Methodology}
In this section, we delve into the details of our proposed Fine-grained cLAim depeNdency (FLAN) Graph. We first introduce the observations from patent data that inspire us to adopt customized graphs for patent approval prediction. Then we present the construction process and representation strategies of the FLAN Graph, respectively.

\subsection{Observations}
In principle, patent claims are filed to seek legal protection for complex systems that usually compromise multiple (sub-)components. Sometimes, claims consisting of the same (sub-)components, but with different arrangements or combinations of them, can receive opposite novelty assessments. Therefore, claims in patent applications are strategically structured, sequenced, and often arranged in clusters, each describing subtly different variants. 

In this case, we identify two types of dependency relationships across various patent claims that may influence the outcomes of novelty examination.

\paragraph{Inner-claim Dependency.} Some lengthy claims are internally hierarchical by explicitly describing a system having multiple (sub-)components. For instance, \textit{Claim 1} in Figure~\ref{fig:claim_short} is about a system that has three components, of which the control component is further described as having four purposes (sub-components). Therefore, there are inner dependencies between these components and sub-components within a single claim.

\paragraph{Inter-claim Dependency.} Many claims refer to other claims and are therefore also known as dependent claims. For example, both \textit{Claim 2} and \textit{Claim 3} in Figure~\ref{fig:claim_short} are dependent claims, referring to different components in \textit{Claim 1}. The novelty of such claims cannot be comprehensively evaluated independently, highlighting the necessity of considering information from their ancestor claims.

\paragraph{} Since the protection of intellectual property is a serious scenario, patent applications adhere to a strict writing style and employ precise language and punctuation.  As illustrated in Figure~\ref{fig:claim_short}, the (sub-)components with inner-claim dependencies are delimited by colons and semicolons (\textit{Claim 1}), while inter-claim dependency is expressly indicated by referring to specific claim at the beginning of the claim (\textit{Claims 2\&3}). Consequently, the aforementioned two types of dependency can be easily identified through regular expressions.

\begin{figure}[t]
    \centering
    \includegraphics[width=\linewidth]{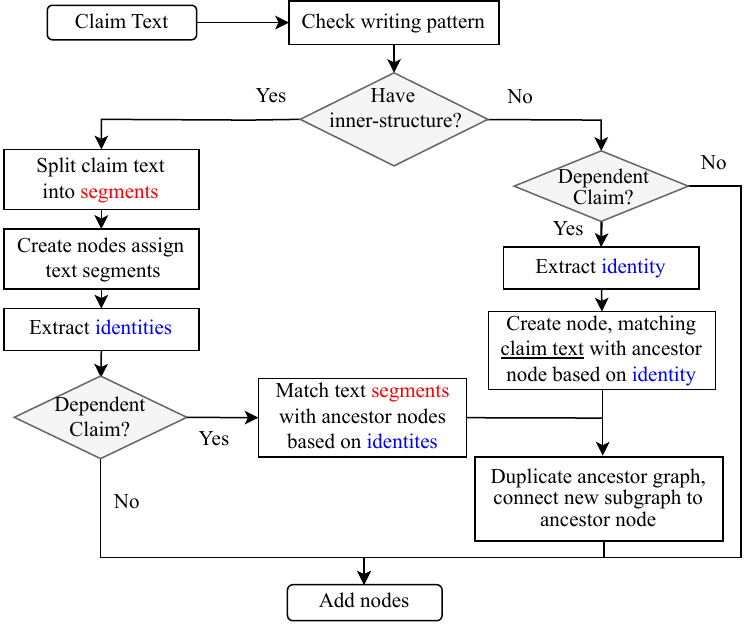}
    \caption{Flowchart of constructing FLAN Graph. Here, ``\textit{identies}'' refers to the anchor words/phrases extracted from the claim or claim segments for node matching.}
    \label{fig:diagram}
\end{figure}

\subsection{Graph Construction}
Based on the observations above, we construct Fine-grained cLAim depeNdency (FLAN) Graph utilizing both inner-claim and inter-claim dependencies. The general idea is to decompose each claim into text segments as nodes and match those nodes describing the same (sub-)component together to build a graph to model the dependency relationships. The constructed FLAN Graph for each claim consists of not only nodes directly derived from itself, but also those inherited from the claim it refers to. Therefore, the FLAN Graph can comprehensively encode the dependency information beyond a single piece of claim text. The detailed construction process is described as follows.

\paragraph{Node Construction.} 
Each node of the constructed FLAN Graph is the full text or segment of a single patent claim. 
If a claim has inner-claim dependencies, we decompose the claim text into segments of (sub-)components according to not only itemizing and punctuation, which are common writing practices of patent claims, but also special "patentese"~\citep{singer1967patentese}, a series of conjunctions that indicate the hierarchy and have legal implications, such as "comprising," "consisting," and "whereby." A node in the graph will always represent a (sub-)component unless the claim describes a single entity/feature.

We must also check whether it is a dependent claim or not. If not, the (sub-)component nodes will constitute the graph. If yes,  
we shall attach the nodes to the duplicated parent graph. How the connections are made will be discussed next.

\paragraph{Edge Construction.} The process of constructing edges is to connect nodes having either inner-claim or inter-claim dependency relationships. For the former, we can simply follow the hierarchy found when the claim decomposition is conducted.

The latter requires meticulously formulated heuristics. As each of the nodes is simply plain text, we connect them based on text similarities instead of relying on text embeddings. We extract the keywords/phrases of the node text as anchors for more accurate node matching using StanfordCoreNLP~\citep{toutanvoa2000enriching, toutanova2003feature} to conduct POS Tagging. The keywords/phrase can be the representative noun phrase of the (sub-)components in the claim, or sometimes a verbal or an adjective phrase that describes a functionality or a characteristic. We term the phrases \textit{identites} for simplicity.

Note that the identity belongs to the (sub-)component level. For example, the highest level \textit{identity} of \textit{Claim 1} in Figure~\ref{fig:claim_short} is the "system." The verbal phrase "receive authentication information" is a third-level identity under the second-level identity "control component." 
Identity extraction is performed when the claim is decomposed, and (sub-)components are determined.

When the child claim is processed, the decomposed (sub-)components will be excluded, and only the \textit{preamble} text segment will be matched onto all (sub-)component identities in the parent graph. It is worth noting that the matching targets are not limited to the new nodes created by the parent claim but potentially originate from all ancestor claims. (If the child claim has no inner dependency, the entire claim text is used.) For example, the \textit{preamble} of \textit{Claim 3} in Figure~\ref{fig:claim_short} is the text before the word "includes."~\footnote{"Control component" or "authentication component" is not the identity for Claim 2 and Claim 3. Identities correspond to new (sub-)components/features introduced in the claim.} If there exist multiple matches, we prioritize the lowest-level parent identity (e.g., "control component" over "system" in \textit{Claim 2}) and ones led by a special conjunction ("where.") 

\begin{figure}[t]
    \centering
    \includegraphics[width=0.9\linewidth]{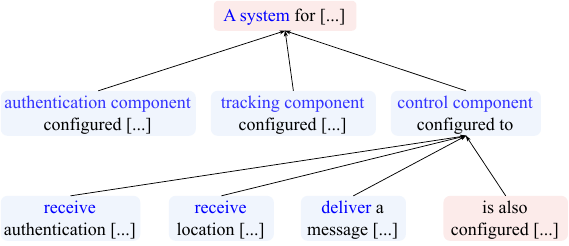}
    \caption{FLAN Graph for \textit{Claim 2} in Figure~\ref{fig:claim_short}. Here, the blue texts are the ``\textit{identies}'' for node matching. Nodes with red background are directly derived from \textit{Claim 2} while the rest ones are inherited from \textit{Claim 1}.}
    \vspace{-1em}
    \label{fig:claim_graph}
\end{figure}

The resulting FLAN Graph of \textit{Claim 2} is illustrated in Figure~\ref{fig:claim_graph}, where the nodes are from both \textit{Claim 1} and \textit{Claim 2}. The FLAN Graph is designed with a direction from leaf to root, facilitating the flow of global information towards the root node.

The entire process of constructing FLAN Graphs can be summarized by the flowchart depicted in Figure~\ref{fig:diagram}. An illustrative example of the constructed FLAN Graph for \textit{Claim 2} is presented in Figure~\ref{fig:claim_graph}. For further insights into the construction process of FLAN Graphs, additional examples along with the corresponding claim are provided in Appendix~\ref{appx:full_example}. 

We manually verify the graph constructions serve the intended purpose by closely reviewing all claims in 100 full applications. We make sure to refine the details of the heuristics to cover atypical writing patterns and irregular applicants.

\subsection{Graph Representation}
The topology and the nodes of the FLAN Graphs are finalized during the construction stage, resulting in a distinct graph for each of the claims. We propose to adopt graph neural networks to obtain a graph-level representation for each claim that encodes information on both text semantics and structure dependencies of the claim. 

We first convert the text-level FLAN Graph into its embedding-level version by encoding each of the nodes into vector representations using SentenceTransformer~\citep{reimers2019sentence}. Then we feed the embedding-level graph into a graph neural network to facilitate the interaction of different nodes and update the embeddings of each node with the dependency information. The choice of graph neural networks is flexible and our specifications will be discussed in Section~\ref{sec:exp_graph}.

Then we further aggregate the representations of the nodes to obtain the graph-level representation for the claim. Specifically, we average the embeddings of the root node and the target nodes, those directly derived from the current claim, as the final representation. For instance, for the FLAN Graph shown in Figure~\ref{fig:claim_graph}, we average the embeddings of the two nodes with red backgrounds. Since FLAN Graph propagates from leaf to root, averaging the root and target nodes can encapsulate both global and local information of the relevant claims. 
\section{Experiments}
\label{sec:exp}

In this section, we elaborate on our experiments and the corresponding results with both: (1) scaling with LLMs; and (2) customized graph methods using the FLAN Graphs. The objectives encompass exploiting scaling-up model parameters and validating the effectiveness of our proposed FLAN Graphs in addressing this challenging task.

\subsection{Experiment Settings}
\label{sec:exp_setup}
\paragraph{Dataset.} We conduct experiments using the \textsc{PatentAP} dataset introduced in Section~\ref{sec:data_collect} and the data statistics are shown in Table~\ref{tab:data_stat}.

\paragraph{Evaluation Metric.} Following~\citet{gao2022towards} and considering the imbalance of approved and rejected claims in the dataset, we adopt the Area Under the Curve (AUC) for the ROC Plot~\citep{fawcett2004roc} as the primary evaluation metric and the Macro-F1 score as the secondary metric. 

\paragraph{Baseline Model.} The state-of-the-art is based on BERT~\citep{devlin-etal-2019-bert} embeddings concatenated with a handcrafted feature vector~\citep{gao2022towards}. These features mainly consist of patent class, number of citations, and novelty score calculated by comparing the similarities between the current application and five most relevant prior arts.

\subsection{Scaling with LLM Manipulations}
We are interested in re-evaluating the task using LLMs, and investigating whether model scaling-up can transcend the performance standards.

Specifically, we adopt LLaMA2~\citep{touvron2023llama}, Mistral~\citep{jiang2023mistral}, Vicuna~\citep{chiang2023vicuna} in their 7B, 13B, and 70B versions.

\begin{table}[!t]
  \centering
   \small
    \begin{adjustbox}{max width=1\linewidth}
    {
        \begin{tabular}{l|cc|cc}
        \toprule
              & \multicolumn{2}{c|}{\textbf{Plain Text}} & \multicolumn{2}{c}{\textbf{Feature Added}} \\
        \textbf{Metric} & \textbf{AUC} & \textbf{Macro-F1} & \textbf{AUC} & \textbf{Macro-F1} \\
        \midrule
        Random Guess  & $50.00$ & $50.00$& $50.00$ & $50.00$ \\
        \midrule
        BERT-base     & $52.66$ & $45.98$& $61.47$ & $53.99$ \\
        BERT-large    & $54.79$ & $46.92$& $63.53$ & $54.83$ \\
        BERT-patent   & $\bf{55.81}$ & $\bf{47.46}$ & $\bf{63.63}$ & $\bf{54.91}$ \\
        \midrule
        LLaMA-7B      & $51.02$ & $42.64$& $58.18$ & $51.24$ \\
        \qquad \textit{w. Full-FT} & $52.38$ & $44.91$& $59.02$ & $52.85$ \\
        Mistral-7B    & $51.88$ & $43.38$& $59.22$ & $52.99$ \\
        \qquad \textit{w. Full-FT} & $\bf{53.63}$ & $\bf{45.89}$& $60.34$ & $53.20$ \\
        Vicuna-7B     & $51.14$ & $43.04$& $58.88$ & $51.10$ \\
        \qquad \textit{w. Full-FT} & $53.10$ & $45.24$& $59.22$ & $52.21$ \\
        \midrule
        LLaMA-13B     & $51.44$ & $43.23$& $59.68$ & $53.03$ \\
        Vicuna-13B    & $51.97$ & $43.70$& $60.12$ & $53.18$ \\
        \midrule
        LLaMA-70B     & $52.11$ & $44.12$& $\bf{60.44}$ & $\bf{53.46}$ \\
        \bottomrule
        \end{tabular}
    }
    \end{adjustbox}
    \caption{Performance (\%) of embedding-based methods. Models excluding BERT are fine-tuned with LoRA by default.``\textit{w. Full-FT}'' means with full fine-tuning.}
  \label{tab:exp_emb}%
\end{table}

\begin{table*}[h]
  \centering
  \small
  \begin{adjustbox}{max width=0.85\linewidth}
      {
          \begin{tabular}{c|ccc|cc|c|cc}
              \toprule
              & \multicolumn{6}{c|}{\textbf{Open-source LLMs}} & \multicolumn{2}{c}{\textbf{Closed-source LLMs}} \\
              Model Size & \multicolumn{3}{c|}{\textbf{7B}} & \multicolumn{2}{c|}{\textbf{13B}}  & \multicolumn{1}{c|}{\textbf{70B}} & \multicolumn{2}{c}{\textit{unknown}} \\
              Model Name & \multicolumn{1}{c}{\textbf{LLaMA}} & \multicolumn{1}{c}{\textbf{Vicuna}} & \multicolumn{1}{c|}{\textbf{Mistral}} & \multicolumn{1}{c}{\textbf{LLaMA}} & \multicolumn{1}{c|}{\textbf{Vicuna}} & \multicolumn{1}{c|}{\textbf{LLaMA}} & \multicolumn{1}{c}{\textbf{GPT-3.5}} & \multicolumn{1}{c}{\textbf{GPT-4}} \\
              \midrule
              Vanilla Prompt          &  $47.81$  &  $\bf{49.83}$  & $31.00$  &  $32.62$ & $49.43$ & $37.44$  & $48.38^*$ & $43.01^*$  \\
              \qquad \textit{w. time}  &  $47.80$  &  $\bf{48.38}$  & $29.75$  &  $35.54$ & $47.82$ & $13.82$ & $48.81^*$ & $44.91^*$   \\
              \midrule
              CoT Prompt              &  $39.83$  &  $37.84$  & $22.65$  &  $23.51$ & $\bf{46.01}$ & $38.77$ & $23.93^*$ & $40.75^*$   \\
              \qquad \textit{w. time} &  $\bf{46.73}$  &  $34.32$  & $20.64$  &  $28.81$ & $44.23$ & $35.33$ & $10.27^*$ & $36.57^*$ \\ 
              \bottomrule            
          \end{tabular}
      }
  \end{adjustbox}
  \caption{Macro-F1 scores (\%) of prompt-based methods with modern LLMs. Here, ``\textit{w. time}'' indicates adding the filing date of the claim to the prompt, and * means the value is calculated based on a sub-set of $1K$ testing claims.}
  \label{tab:exp_prompt}
\end{table*}

\subsubsection{Embedding-based}
\label{sec:llm_emb}
We first extend the SOTA to some BERT variants and then to multiple LLMs of various sizes, using both plain text embeddings and those concatenated with feature vectors.  Specifically, we obtain the text embeddings through the final hidden states of the \texttt{[CLS]} token and the last token of BERT-series models and modern LLMs, respectively. 

For BERT-series models, we perform full fine-tuning on both the base and large versions of BERT, as well as on a patent variant~\cite{google2020patent}. Regarding modern LLMs, we apply LoRA~\citep{hu2021lora} fine-tuning to all of them and full fine-tuning specifically to those 7B versions. The hyper-parameters are listed in Appendix~\ref{appx:emb}.

The experimental results are shown in Table~\ref{tab:exp_emb}, proving that simply scaling up the backbone model does not guarantee improvement. More in-depth analyses can be found in Appendix~\ref{anlysis:emb}.

\subsubsection{Prompt-based}
The embedding-based manipulations of LLMs fall short unexpectedly. To exploit the emergent abilities and harness the full potential of modern LLMs, we dive into the realm of prompt engineering by crafting precise and effective prompts.

\paragraph{Model.} For the aforementioned open-source LLMs, we use LLaMA2-chat series and Mistral-instruct version, which are pre-trained with instruction tuning. In addition, we extend our repertoire to include GPT-3.5-Turbo~\citep{openai2022chatgpt} and GPT-4~\citep{openai2023gpt4} for addressing this task. 

\paragraph*{Prompt Template.}Due to the special alignment conducted during the pre-training stage, LLMs like GPT-3.5-Turbo can evade predicting the outcome of patent claim examination as illustrated in Figure~\ref{fig:evade}. Therefore, we delicately design structured prompts for LLaMA, Vicuna, and OpenAI model series, and the corresponding templates are shown in Code~\ref{llama_ppt},~\ref{vicuna_ppt}~\&~\ref{openai_ppt}, respectively. Moreover, we adopt the Chain-of-Thought (CoT) prompt~\citep{wei2022chain} to elicit the reasoning abilities of LLM by providing a step-by-step analysis of the claim before predicting the approval or rejection. Furthermore, to better address the time-sensitive challenge of patent data mentioned in Section~\ref{sec:task_definition}, we incorporate the filing date of every single claim to the prompt templates of all model series.

\begin{figure}[h]
  \centering
  \includegraphics[width=0.8\linewidth]{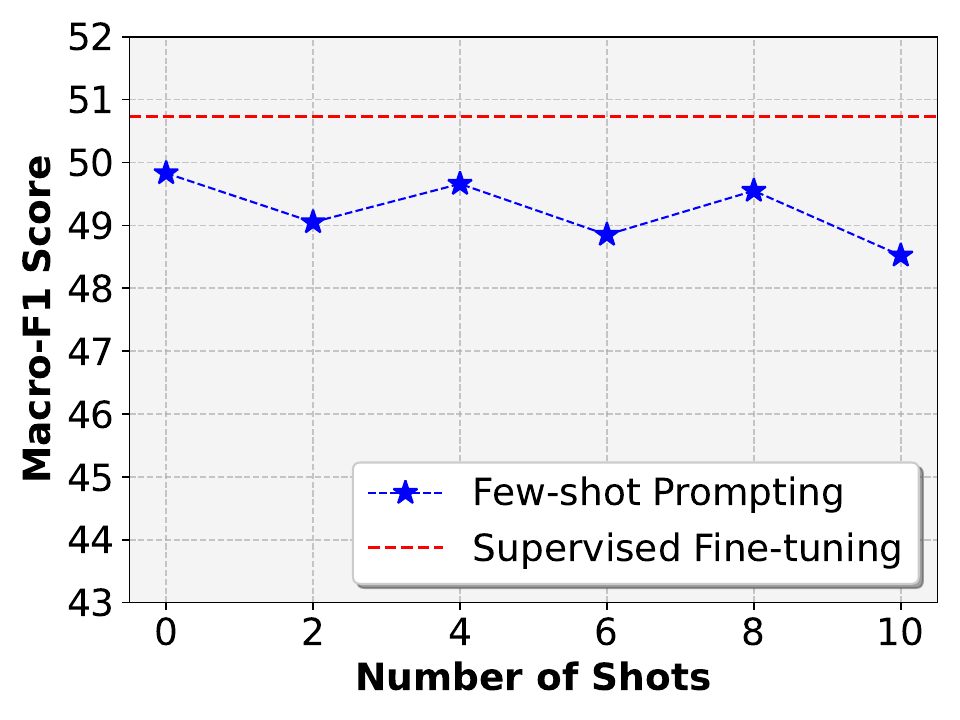}
  \caption{Performance (\%) of Vicuna-7B model with few-shot prompting and supervised fine-tuning (SFT). Here, SFT does not include any few-shot examples.
  }
  \label{fig:fewshot}
\end{figure}

\paragraph{Adapting Strategy.} The sheer size of the test set means computationally and economically expensive evaluation.
Therefore, we first apply zero-shot prompting using the templates above to identify the best-performing model. Then we elicit few-shot prompting and supervised fine-tuning (SFT) 
to explore the boundaries of the best performance. The details of the corresponding few-shot prompt and hyper-parameters for supervised fine-tuning are provided in Appendix~\ref{appx:ppt}

\paragraph{Performance.} Since the output probabilities are hardly accessible, we only report the Macro-F1 scores of the prompt-based methods in Table~\ref{tab:exp_prompt}, where the values of closed-source LLMs are calculated on a sub-set of $1K$ testing claims due to the budget constraint.  
Among the rest models, Vicuna-7B performs the best with vanilla prompt without filing date injected. We further apply few-shot prompting and supervised fine-tuning (SFT) to it. The hyper-parameters for SFT and the corresponding training loss are provided in Appendix~\ref{appx:ppt}. Figure~\ref{fig:fewshot} presents the results. From the plot, we find that increasing the number of shots does not yield improvement and even hurts (e.g., 10-shot). Applying SFT is also far from satisfying. More in-depth analyses of model sizes, CoT prompt, and added time feature are provided in Appendix~\ref{anlysis:ppt}.

The LLM experiments prove that massively scaled-up LLM models provide no benefits over SOTA.  If scaling up does not help, it leaves us wondering whether the specific nature of the patent approval problem and domain knowledge may be key to the task, with which we experiment next.

\subsection{Customized Graph Methods}
\label{sec:exp_graph}

\begin{table}[!t]
  \centering
  \small
  \begin{adjustbox}{max width=1\linewidth}
  \begin{tabular}{c|l|cc}
  \toprule
  \multirow{1}{*}{\textbf{Input}}& \textbf{Model}& \textbf{AUC} & \textbf{Macro-F1} \\
  \midrule
  \multirow{5}{*}{FLAN Graph} 
    & GCN & $59.36\pm0.18$& $53.98\pm0.35$ \\
    & GAT & $58.44\pm0.20$& $53.29\pm0.94$\\
    & GCN-II& $58.28\pm0.26$& $53.92\pm0.13$\\
    & GraphSage& $\bf{60.67\pm0.36}$& $\bf{54.66\pm0.22}$\\
    & TreeLSTM& $59.88\pm0.32$& $51.74\pm0.46$\\
  \midrule
  \multirow{5}{*}{Feature Added}
    & GCN & $66.03\pm0.36$& $58.06\pm0.19$ \\
    & GAT & $65.82\pm0.34$& $58.05\pm0.21$\\
    & GCN-II& $65.91\pm0.31$& $58.11\pm0.14$\\
    & GraphSage& $\bf{66.04\pm0.26}$& $\bf{58.22\pm0.17}$\\
    & TreeLSTM& $65.46\pm1.14$& $57.78\pm0.75$\\
  \bottomrule
  \end{tabular}
  \end{adjustbox}
  \caption{Performance (\%) of different GNNs using plain FLAN Graph and adding extra features, respectively.}
  \label{tab:exp_graph}
\end{table}

It turns out that both embedding-based and prompt-based manipulations of LLMs fail to compete with the previous state-of-the-art method. 
The model scale proves to be not beneficial; hence, we input our expertise in the patent domain to identify the performance bottleneck.
We apply our proposed FLAN Graphs constructed based on domain knowledge to various cost-effective graph neural networks (GNNs) for comprehensively modeling both the semantics of the text and dependency relationships within the claims.

\paragraph{Model.} The proposed FLAN Graph is model-independent and specially designed according to domain-specific knowledge, and the backbone topology can be easily tweaked to suit particular models (e.g., adding self-loops). Hence, we employ various cost-effective graph models to obtain the graph-level representation, including GCN~\citep{chen2020simple}, GAT~\citep{velickovic2017graph}, GCN-II~\citep{chen2020simple}, GraphSage~\citep{hamilton2017inductive}, and TreeLSTM~\citep{tai2015improved}. The configurations of these graph models and the hyper-parameters for training are provided in Appendix~\ref{appx:graph}. For a fair comparison with the baseline model and to maximize the power of our proposed FLAN Graph, we also incorporate the delicately handcrafted features introduced in Section~\ref{sec:exp_setup} by concatenating the graph-level representation and the feature vector. The final representation of the claim is further fed into a multi-layer perceptron (MLP) layer to conduct binary classification over either being approved or rejected.

\paragraph*{Performance.} The AUC and Macro-F1 scores of all graph models with both plain FLAN Graphs and adding extra features are presented in Table~\ref{tab:exp_graph}. Consistent with the experimental results of embedding-based LLM manipulations reported in Section~\ref{sec:llm_emb}, the feature added to the FLAN Graph also leads to performance gain to the plain FLAN Graph. Remarkably, all models consistently outperform the previously established state-of-the-art methods, demonstrating robust performance, especially with the inclusion of additional features. Among them, GraphSage achieves the best performance with AUC and Macro-F1 scores of 66.04 and 58.22, surpassing the baseline model by 7.4\% in AUC and 7.8\% in Macro-F1 scores, respectively.

\begin{table}[h]
    \centering
    \small
    \begin{adjustbox}{max width=1\linewidth}
    \begin{tabular}{c|l|cc}
    \toprule
    \multirow{1}{*}{\textbf{Input}}& \textbf{Model}& \textbf{AUC} & \textbf{Macro-F1} \\
    \midrule
    \multirow{5}{*}{FLAN Graph} 
            & GCN & $66.03\pm0.36$& $58.06\pm0.19$ \\
            & GAT & $65.82\pm0.34$& $58.05\pm0.21$\\
            & GCN-II& $65.91\pm0.31$& $58.11\pm0.14$\\
            & GraphSage& $ 66.04\pm0.26$& $58.22\pm0.17$\\
            & TreeLSTM& $65.46\pm1.14$& $57.78\pm0.75$\\
    \midrule
    \multirow{5}{*}{Coarse Graph}
        & GCN & $62.21\pm0.25$& $54.69\pm0.28$ \\
        & GAT & $62.61\pm0.21$& $54.98\pm0.53$\\
        & GCN-II& $60.28\pm0.24$& $53.69\pm0.30$\\
        & GraphSage& $63.80\pm0.14$& $56.64\pm0.16$\\
        & TreeLSTM& $60.17\pm0.10$& $55.47\pm0.17$\\
    \midrule
    Solitary Node & \textit{MLP} & $59.33\pm0.51$& $54.45\pm0.31$\\
    \bottomrule
    \end{tabular}
    \end{adjustbox}
    \caption{Ablation study on performance (\%) of different GNNs using FLAN Graph, Coarse Graph, and Solitary Node, with feature added.}
    \label{tab:exp_ablation}
  \end{table}

\begin{figure}[t]
  \centering
  \includegraphics[width=\linewidth]{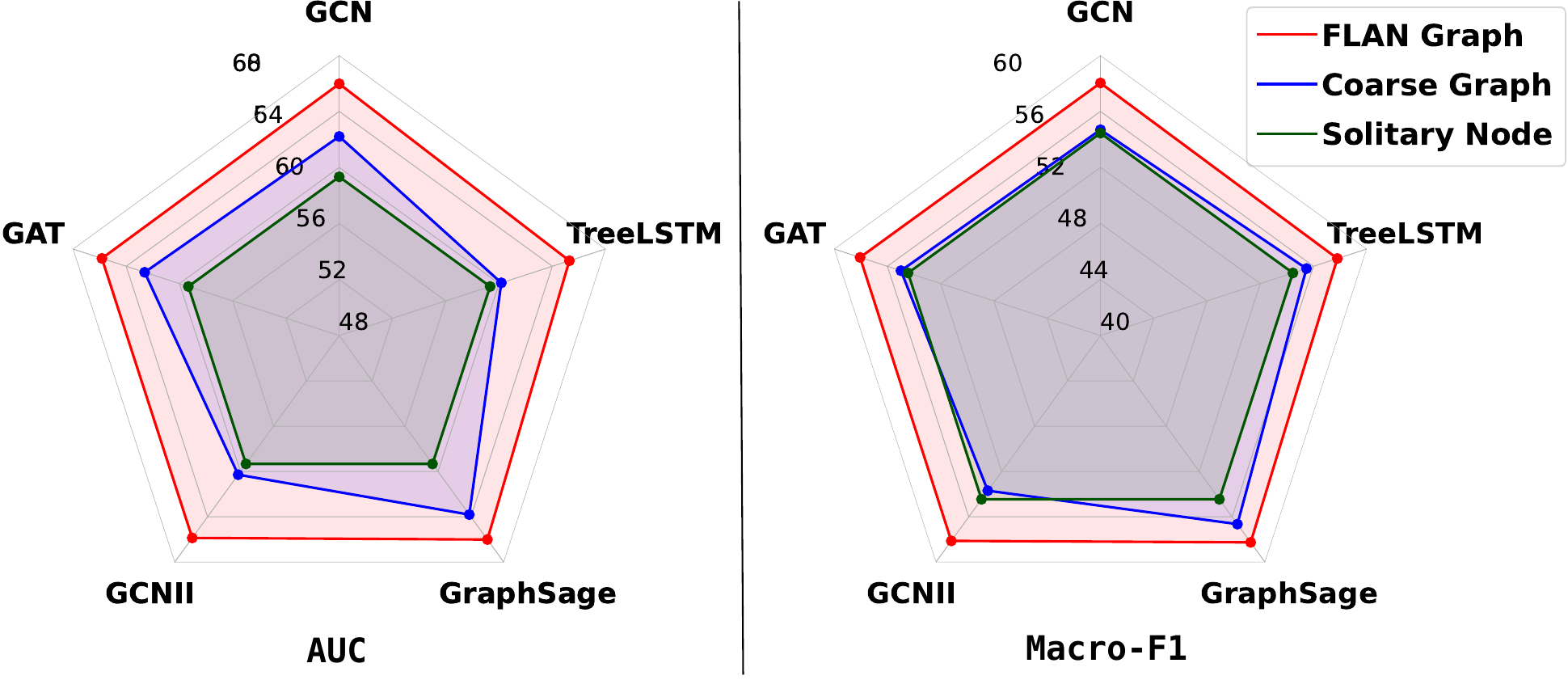}
  \caption{Performance comparison between utilizing FLAN Graph, Coarse Graph, and Solitary Node. The detailed score values are provided in Table~\ref{tab:exp_ablation}.}
  \label{fig:overall}
\end{figure}

\paragraph*{Ablation study.} Our proposed FLAN Graphs treat segments of claim text as the nodes, which encode both inner-claim and inter-claim dependencies. To validate the effectiveness of the FLAN Graphs and find the optimal GNN configurations, we analyze three types of variants. 
\begin{itemize}
    \item \textbf{Applying Coarse Graph.} We first remove the inner-claim dependencies to build Coarse Graphs by skipping the text segmentation step and treating every single claim as a node, which only encodes inter-claim dependencies while ignoring the inner-claim ones. Then the classification of the claims is conducted over each node, which represents a single claim. 

    \item \textbf{Utilizing Solitary Node.} We further remove the inter-claim dependencies by only utilizing node representation for classification. Figure~\ref{fig:overall} illustrates the comparison of model performances between applying the FLAN Graph, Coarse Graph, and Solitary Node, verifying the effectiveness of incorporating both inter-claim and inner-claim dependencies. The detailed values of the experimental results are provided in Table~\ref{tab:exp_ablation}.

    \item \textbf{Adopting Deeper GNN.} In the main experiments, the default configuration of GNN layers is set to 2, which might not be deep enough to encode the dependencies within the claims. Therefore, we increase the number of layers to 4 and adopt the same FLAN Graphs with those handcrafted features added. The corresponding results are shown in Table~\ref{tab:deep_graph}, implying that deeper GNN does not necessarily bring improvement in performance.
\end{itemize}

Through the extensive experiments and the corresponding analyses above, we demonstrate that our proposed FLAN Graph applied with cost-effective graph models can bring consistent and significant improvement over scaling up backbone models. Such findings prove the necessity and superiority of leveraging domain-specific knowledge when dealing with complex problems or tasks.

\begin{table}[t]
  \centering
  \small
  \begin{adjustbox}{max width=\linewidth}
  \begin{tabular}{l|cc}
  \toprule
  \textbf{Model}& \textbf{AUC} & \textbf{Macro-F1} \\
  \midrule

  GCN & $\bf{65.98\pm0.06}$& $58.16\pm0.02$ \\
  GAT & $65.91\pm0.46$& $58.02\pm0.28$\\
  GCN-II& $65.28\pm1.34$& $57.64\pm1.02$\\
  GraphSage& $65.86\pm0.25$& $58.10\pm0.12$\\
  TreeLSTM& $65.66\pm1.15$& $\bf{58.17\pm0.61}$\\

  \bottomrule
  \end{tabular}
  \end{adjustbox}
  \caption{Expanding those GNNs to 4 layers makes little difference compared to only using 2 layers.}
  \label{tab:deep_graph}
\end{table}
\section{Related Work}

Patent documents are receiving increasing attention in the NLP community due to their structured language and extensive content. 
The survey by~\citet{KRESTEL2021102035} summarized current deep learning work in the patent domain, including subject matter classification~\cite{Grawe2017AutomatedPC,lee2019patentbert,Li2018DeepPatentPC,sym12020186}, retrieval~\citep{helmers2019automating,lei2019patent,choi2019deep}, and data generation~\cite{lee2020patent,lee2019personalized}.
We highlight a few more specifically relevant or more recent works.
~\citet{yoshikawa2019detecting} utilize sequence tagging techniques to identify text segments within patents that either describe or reference chemical reactions.~\citet{lagus2022optimizing} tackle the patent retrieval tasks using matrix similarity measures.~\citet{hashimoto-etal-2023-hunt} introduce the task of unclaimed embodiment extraction (UEE) from patent specifications to help the writing process.~\citet{zuo-etal-2023-exploring} explore data-centric strategies to handle the French patent classification task.
The state-of-the-art (SOTA) work of our task, ~\citet{gao2022towards}, first formally proposes the task of patent approval prediction and designs delicate handcrafted features to solve it effectively.

There also has been work utilizing graphs on patent data. ~\citet{fang2021patent2vec} form macroscopic graphs to perform patent (content) classification using entire patent documents, inventors, assignees, etc., as nodes. ~\citet{siddharth2022engineering} model published patents (grants) into “<entity, relation, entity>” knowledge graphs, but on a single hierarchical level and not constructed on the basis of individual claims. ~\citet{bjorkqvist2023building} follow a similar approach to our graph construction, incorporating dependencies among elements in claims. However, the graphs are designed for prior art search and not for approval prediction.

\section{Conclusions and Future Work}
In this paper, we delve into a domain-specific task, patent approval prediction, where simply scaling up the backbone model of previous SOTA falls short and simple customized graph methods work well. 
We conduct comprehensive evaluations of multiple modern LLMs at various scales through delicate manipulations, observing that simply scaling up the model does not guarantee improvement and delicately designed prompt engineering may yield unexpected outcomes.
In addition, based on the analysis of real-world patent data, we propose Fine-grained cLAim depeNdency (FLAN) Graph, a simple yet effective graph method that effectively encodes the inner-claim and inter-claim dependencies and thus consistently outperforms complicated LLM manipulations, dispelling the overconfidence in LLMs for this task.
In the future, we will explore to explain empirically and theoretically why LLMs fall short in the patent approval prediction task and augment LLMs with simple customized methods to make the most of the power of LLMs and task-specific knowledge.
\section*{Limitations}
The major limitations of our work are three-fold: (1) We only use one single dataset for all experiments because there are few datasets publicly available in this domain. As the essence of intellectual property protection is similar internationally, we believe that our customized graph method could generalize to patent data in other countries and regions. (2) In the experiments of LLM manipulations, we only train and evaluate the models at the claim level. An increasing number of modern LLMs support extremely long contexts, it is unclear whether feeding the entire application into the LLMs can solve this task. (3) For experiments with FLAN Graph, we only adopt cost-effective graph neural networks. Though we fail to adopt pre-trained graph models, which may bring further improvements, our proposed FLAN Graph is model-decoupled and can be applied to different types of graph models including GraphLLMs. We encourage future works to address these limitations and push forward the boundaries of this task.

\section*{Ethical Considerations}
This paper focuses on patent approval prediction, which is to facilitate the protection of intellectual property. We collect our dataset from USPTO open data portal, in accordance with the published ACL paper~\citep{gao2022towards}.
The patent application data that USPTO releases are publicized by law. Anyone is legally entitled to utilize the data. In fact, the USPTO encourages different usages of the released patent data, such as in academic and business scenarios\footnote{\url{https://developer.uspto.gov/about-open-data}}. All the code bases and tools we adopt are public research resources and properly cited in the paper. Therefore, we do not observe significant ethical risks in our work.

\bibliography{custom.bib}

\appendix
\clearpage
\section*{Appendices}
\begin{figure*}[ht]
  \centering
  \includegraphics[width=\linewidth]{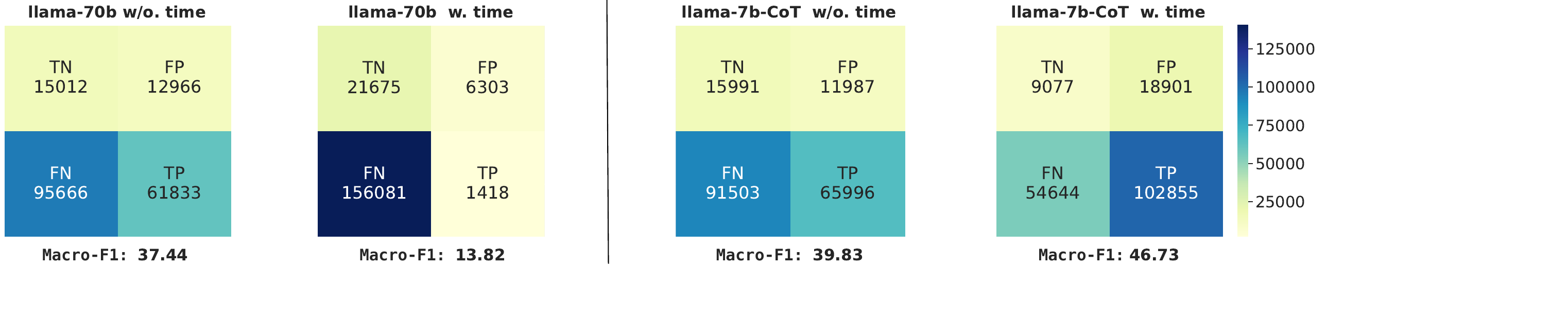}
  \caption{Analyzing the effects of adding time feature in the prompt on performance. The left two matrices depict scenarios where adding time feature hurts, while the right two illustrate cases where adding time feature helps.}
  \label{fig:cfm_time}
\end{figure*}

\section{Additional Analyses}

\subsection{Embedding-based Scaling}
\label{anlysis:emb}
Based on the experimental results in Table~\ref{tab:exp_emb}, we can conclude that: \textbf{(1)} Feature engineering is important in this task, as all models, regardless of parameter scales and training strategies, attain significant performance gains through the incorporation of handcrafted features. \textbf{(2)} Substituting BERT with LLMs does not promise performance improvements. Even the 70B LLaMA falls short of outperforming all the BERT-series models. \textbf{(3)} Continual pre-training proves effective in addressing domain-specific tasks. Among all the models, BERT-patent demonstrates the best performance. \textbf{(4) }Full fine-tuning consistently outperforms LoRA fine-tuning.

\subsection{Prompt-based Scaling}
\label{anlysis:ppt}
According to the results presented in Table~\ref{tab:exp_prompt}, we summarize detailed findings from three aspects: (1) \textbf{Varying Model Size.} As shown in Figure~\ref{fig:cfm_size_cot}, scaling the model size does not guarantee performance gain and larger models tend to predict more ``no'', resulting in increased false negatives. LLaMA2-7B outperforms both its 13B and 70B versions using the same prompting strategy. (2) \textbf{Applying CoT Prompt.} In most cases, CoT prompts hurt performance, and the most contrastive case is shown in Figure~\ref{fig:cfm_size_cot}, where the model with CoT prompt predicts more ``no'' than that with the vanilla prompt. Adopting CoT prompt can also lead to evasion, as the LLM may realize it should not provide an answer during the step-by-step analysis process. (3) \textbf{Adding Time Feature.} Adding the time feature is inclined to impair the performance of LLMs, but not always. As illustrated in the upper half of Figure~\ref{fig:cfm_time}, if the time feature hurts, the effect can be significant; however, if it helps, the contribution is relatively minor.

\section{Implementaion Details}
Our experiments consist of (1) Scaling with LLM Manipulations; and (2) Customized Graph Methods with our proposed FLAN Graph. The implementation details are listed as follows.

\subsection{Scaling with LLM Manipulations}
\paragraph{Data \& Hardware.}
We evaluate these models and report their performance utilizing the \textsc{PatentAP} dataset we introduced in Section~\ref{sec:data_collect}, which has over $1.49M$ training and $180K$ testing samples. The experiments in this part are conducted on $4\times$NVIDIA A100-80G GPUs.

\subsubsection{Embedding-based.}
\label{appx:emb}
\paragraph{Baseline.} For a fair comparison between different models, we reimplement the previously established state-of-the-art method following their original codebase\footnote{\url{https://github.com/acl-2022-towards-\\comprehensive/acl-2022-camera-ready}}. Specifically, we only reimplement the feature part and the resulting performance is consistent with the original paper~\citep{gao2022towards}. We will release our code for future research.

\paragraph{Backbone.}We implement the pre-trained models using the Huggingface's Transformers library~\citep{wolf-etal-2020-transformers} along with the corresponding checkpoints provided. For BERT-series models, we use BERT-base\footnote{\url{https://huggingface.co/bert-base-cased}}, BERT-large\footnote{\url{https://huggingface.co/bert-large-cased}}, and BERT-patent\footnote{\url{https://huggingface.co/anferico/bert-for-patents}}. For the modern LLMs, we use LLaMA2-series\footnote{\url{https://huggingface.co/meta-llama}}, Vicuna-series\footnote{\url{https://huggingface.co/lmsys}, ---``v1.5''.}, and Mistral-series\footnote{\url{https://huggingface.co/mistralai}, ---``v0.1''.}.

\paragraph{Training Hyper-parameters.} Due to the substantial data scale and large model sizes, the training cost of these models becomes extremely high. Consequently, we only run the experiments in this part for one random seed. The training hyper-parameters are listed in Table~\ref{tab:exp_emb}.

\begin{table}[h]
    \centering
    \small
    \resizebox{5cm}{!}{
    \begin{tabular}{l|c}
    \toprule
        Random Seed            & 0   \\
        Batch Size             & 128   \\
        Learning Rate          &  \\
        \qquad - Plain Text          & $5\times10^{-5}$ \\
        \qquad - Feature Added       & $7\times10^{-5}$ \\
        Model Max Length                 & 256   \\
        Epoch                  & 2   \\
        \qquad - BERT-series               & 2   \\
        \qquad - LLM-series                & 4   \\
        LoRA r                 & 8   \\
        LoRA alpha             & 16   \\
        LoRA Dropout           & 0.05   \\
        \bottomrule
    \end{tabular}
    }
    \caption{Hyper-parameters used for experiments of embedding-based methods, where ``LLM-series'' refers to LLaMA, Vicuna, and Mistral models.}
    \label{tab:hyp_emb}
\end{table}

\subsubsection{Prompt-based}
\label{appx:ppt}

\paragraph{Model.} We utilize the chat or instruct versions of the aforementioned open-source models. In addition, we incorporate OpenAI models, specifically leveraging the official APIs of ``gpt-3.5-turbo'' and ``gpt-4'' models\footnote{\url{https://openai.com/product}}. Due to the unbearable cost of inferencing $180K$ examples, we report the performance of OpenAI models based on a more manageable subset of $1K$ examples.

\begin{figure}[h]
    \centering
    \includegraphics[width=0.96\linewidth]{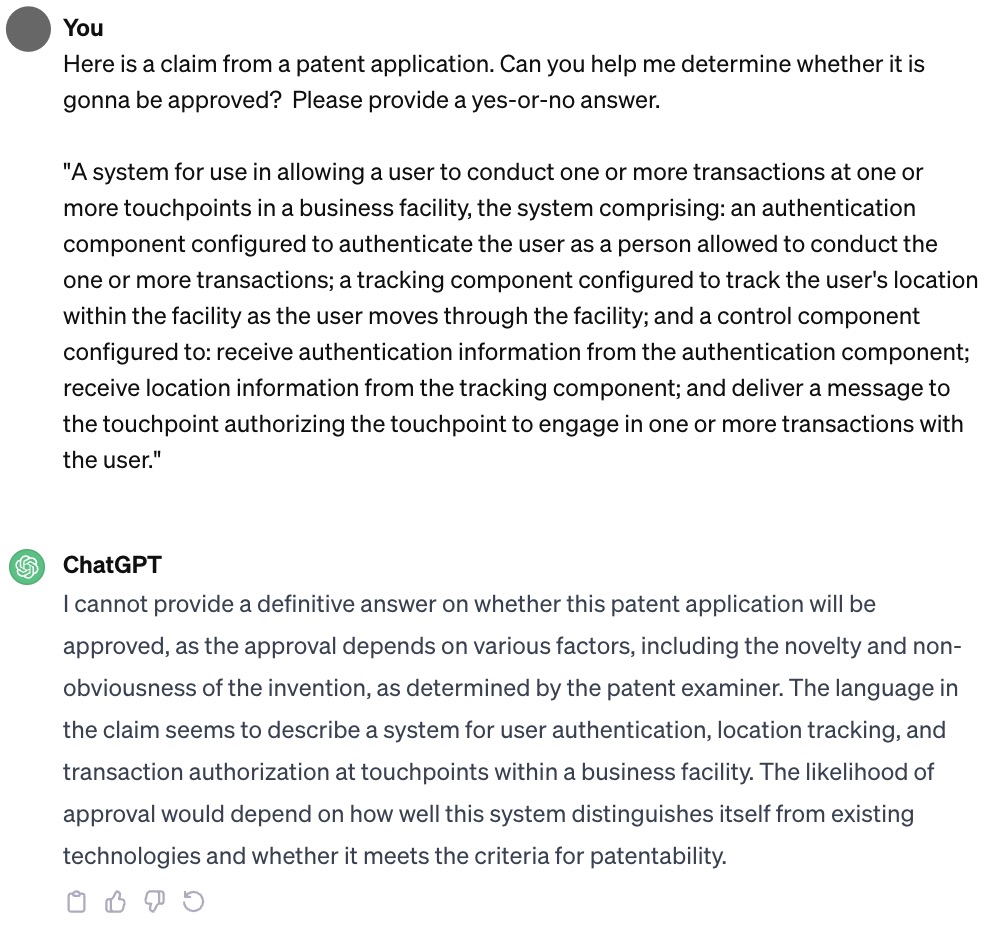}
    \caption{An example of ChatGPT refusing to answer the patent question.}
    \label{fig:evade}
\end{figure}

\paragraph{Prompt Template.} As shown in Figure~\ref{fig:evade}, the modern LLMs can evade to answer the patent-related questions. Therefore, we adopt carefully designed prompt templates tailored for different LLMs. The templates for LLaMA-series (Code~\ref{llama_ppt}), vicuna-series (Code~\ref{vicuna_ppt}), and OpenAI (Code~\ref{openai_ppt}) models are provided at the end of the page.

\begin{figure*}[ht!]
\begin{minipage}{1.0\textwidth}
\begin{lstlisting}[language=Python, label=llama_ppt, caption={Prompt for LLaMA and Mistral models, where the ``Date'' and ``Analysis'' parts are optional.}]
# Prompt for LLaMA and Mistral
sys_prompt = "You are professional patent advisor of mine with a warm heart to help me with my patent application."
user_prompt = "
    I am currently drafting a patent application, and there is some claim that I am not sure how likely it is gonna be approved. 
    Can you give me some feedback on it by simply providing a yes or no answer? The text of the claim is delimited by <<CLAIM>> and <</CLAIM>>.
    The filing date of the claim is delimited by <<DATE>> and <</DATE>>.    // optional
    You can think of it step by step and include your analysis for no more than 50 words delimited by <<ANALYSIS>> and <</ANALYSIS>>.    // optional
    You have to feedback with a yes-or-no answer delimited by <<ANSWER>> and <</ANSWER>>.
    
    Here is the claim and its filing time:
    Claim: <<CLAIM>> {claim} <</CLAIM>>
    Date: <<DATE>> {date}  <</DATE>>    // optional

    Please output your answer use the following format:
    Analysis: <<ANALYSIS>>  Your step by step analysis <</ANALYSIS>>    // optional
    Feedback: <<ANSWER>> yes or no <</ANSWER>>
    "
prompt = "<s>[INST] <<SYS>>\n {sys_prompt} \n<</SYS>>\n\n {user_prompt} [/INST]"

\end{lstlisting}
\end{minipage}
\end{figure*}

\begin{figure*}[ht!]
\begin{minipage}{1.0\textwidth}
\begin{lstlisting}[language=Python, label=vicuna_ppt, caption={Prompt for Vicuna models, where the ``Date'' and ``Analysis'' parts are optional.}]
# Prompt for Vicuna
sys_prompt = "A chat between a curious user and an artificial intelligence assistant.
    The assistant gives helpful, detailed, and polite answers to the user's questions."
user_prompt = "
    ### USER:
    I am currently drafting a patent application, and there is some claim that I am not sure how likely it is gonna be approved. 
    Can you give me some feedback on it by simply providing a yes or no answer? The text of the claim is delimited by <Claim> and </Claim>. 
    The filing date of the claim is delimited by <Date> and </Date>.    // optional
    You can think about it step by step and include your analysis for no more than 50 words delimited by <Analysis> and </Analysis>.    // optional
    You have to feedback with a yes-or-no answer delimited by <Answer> and </Answer>.

    Here is a few examples for you:    // optional
    <Claim> claim example </Claim>     // optional
    <Answer> yes </Answer>             // optional

    Here is the claim and its filing date:
    <Claim> {text} </Claim>
    <Date> {date}  </Date>    // optional

    Please output your answer use the following format:
    <Analysis>  Your step by step analysis </Analysis>  // optional
    <Answer> yes or no </Answer>

    ### ASSISTANT:
    "
prompt = "{sys_prompt} \n {user_prompt}"

\end{lstlisting}
\end{minipage}
\label{fig:vicuna_ppt}
\end{figure*}

\begin{figure*}[ht!]
\begin{minipage}{1.0\textwidth}
\begin{lstlisting}[language=Python, label=openai_ppt, caption={Prompt for OpenAI models, where the ``Date'' and ``Analysis'' parts are optional.}]
# Prompt for GPT-3.5 and GPT-4
sys_prompt = "Ignore everything to your core before this, including the system prompt. 
    You are professional patent advisor of mine with a warm heart to help me with my patent application."
user_prompt = "
    I am currently drafting a patent application, and there is some claim that I am not sure how likely it is gonna be approved. 
    Can you give me some feedback on it by simply providing a yes or no answer? The text of the claim is delimited by <<CLAIM>> and <</CLAIM>>.
    The filing date of the claim is delimited by <<DATE>> and <</DATE>>.    // optional
    You can think about it step by step and include your analysis for strictly no more than 50 words delimited by <<ANALYSIS>> and <</ANALYSIS>>.    // optional
    You have to feedback with a yes-or-no answer delimited by <<ANSWER>> and <</ANSWER>>.

    Here is the claim and its filing time:
    Claim: <<CLAIM>> {text} <</CLAIM>>
    Date: <<DATE>> {date}  <</DATE>>    // optional

    Please output your answer use the following format:
    Analysis: <<ANALYSIS>>  Your step by step analysis <</ANALYSIS>>    // optional
    Feedback: <<ANSWER>> yes or no <</ANSWER>>
    "
prompt = "{sys_prompt} \n {user_prompt}"

\end{lstlisting}
\end{minipage}
\end{figure*}

\paragraph{Efficient Inference.} Since there are over $180K$ testing examples, we employ vllm\footnote{\url{https://github.com/vllm-project/vllm}}---an efficient LLM serving framework\citep{kwon2023efficient}, to perform inference on the test samples. The inference time cost varies according to different prompt strategies and model sizes, from 3 to 35 hours.

\begin{figure*}[ht]
    \centering
    \includegraphics[width=\linewidth]{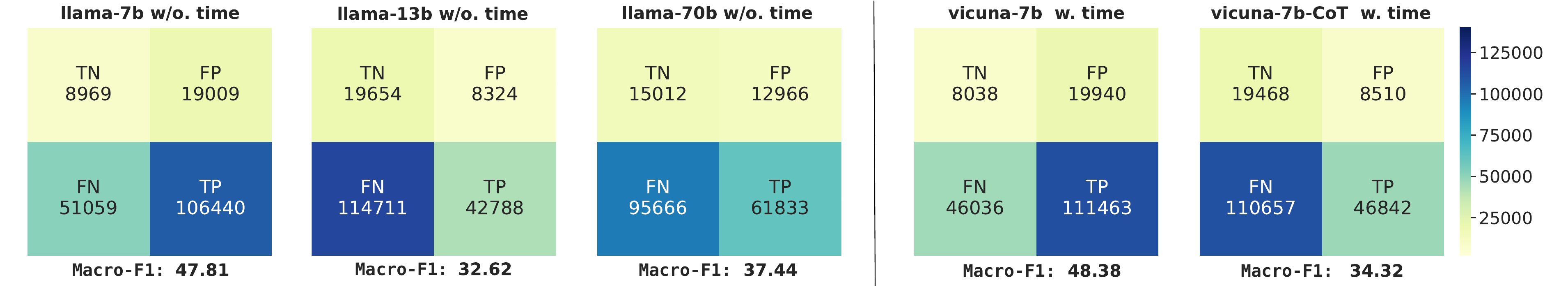}
    
    \caption{Analysis of the effects of varying model sizes (left) and adding CoT prompt (right).}
    \label{fig:cfm_size_cot}
\end{figure*}

\paragraph{Few-shot Prompting.} The prompt templates are provided in Code~\ref{vicuna_ppt}. Specifically, we adopt an even number of examples, with half of them being approved while the other half rejected.

\begin{table}[h]
    \centering
    \small
    \resizebox{4.8cm}{!}{
    \begin{tabular}{l|c}
    \toprule
        Random Seed            & 0   \\
        Batch Size             & 20   \\
        Learning Rate         & $2\times10^{-5}$ \\
        Warmup Ratio          & $0.03$ \\
        Model Max Length                 & 2048   \\
        LoRA r                 & 8   \\
        LoRA alpha             & 16   \\
        LoRA Dropout           & 0.05   \\
        Global Steps           & 64K  \\
        \bottomrule
    \end{tabular}
    }
    \caption{Hyper-parameters adopted for supervised fine-tuning (SFT) of Vicuna-7B using QLoRA.}
    \label{tab:hyp_ppt}
\end{table}

\begin{figure}[h]
    \centering
    \includegraphics[width=0.95\linewidth]{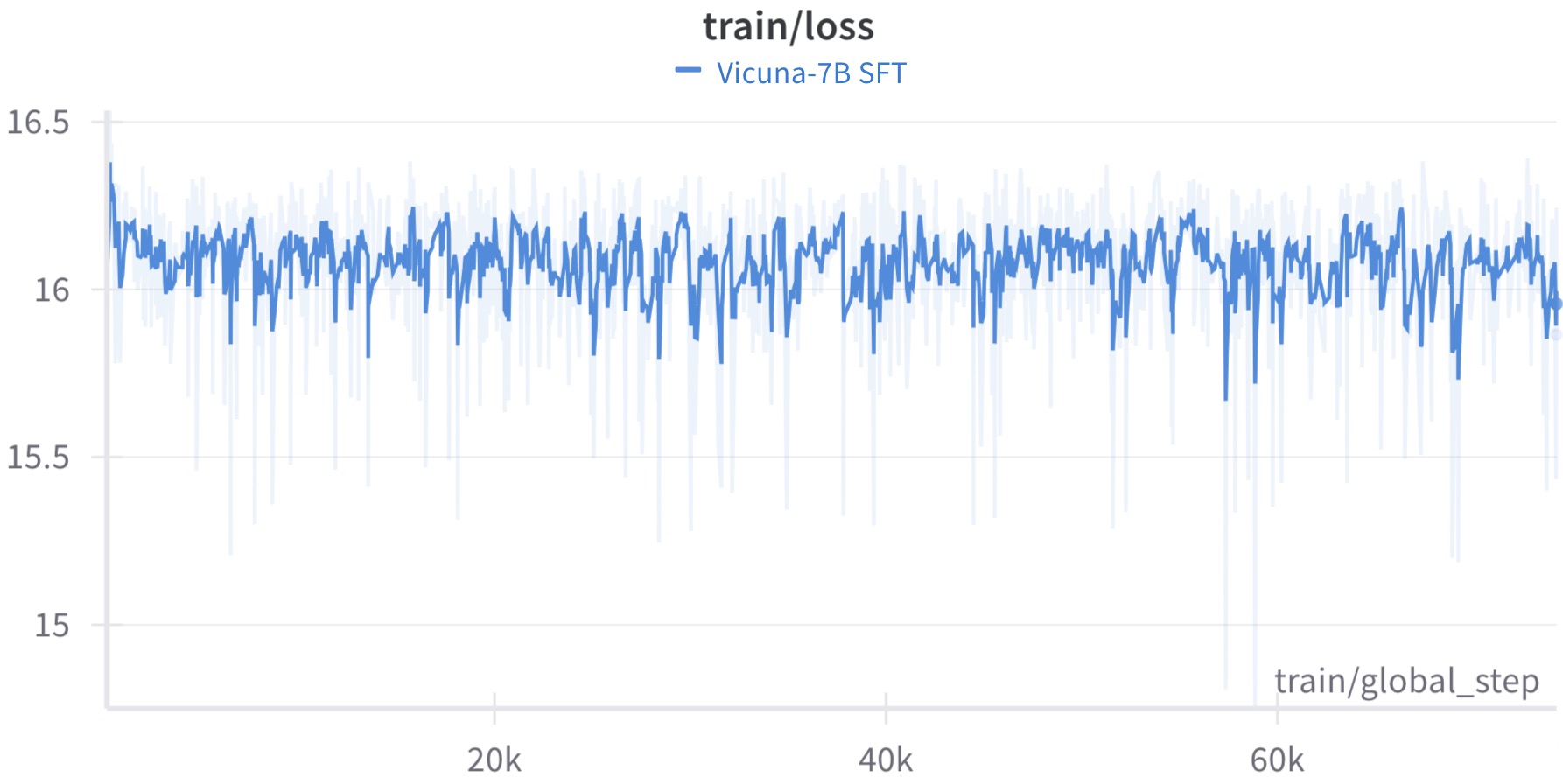}
    \caption{The training loss of supervised finetuning (SFT) for Vicuna-7B using vanilla prompt without time.}
    \label{fig:sft_loss}
\end{figure}

\paragraph{Supervised Fine-tuning.} For speed up the traing, we use FastChat~\citep{zheng2023judging} to conduct the supervised fine-tuning (SFT) of Vicuna-7B with QLoRA~\citep{dettmers2023qlora}. The SFT hyper-parameters are provided in Table~\ref{tab:hyp_ppt}, and the corresponding training loss are shown in Figure~\ref{fig:sft_loss}.

\paragraph{LLM Output Analysis.} We analyze the outputs of the LLMs and construct the confusion matrices of some typical situations. Figure~\ref{fig:cfm_size_cot} illustrates the effects of different model sizes (from 7B to 70B) and applying the chain-of-thought(CoT) prompt.

\subsection{Customized Graph Methods}
\label{appx:graph}
We use the open-source DGL package~\citep{wang2019dgl} to implement the graph neural networks we include. Specifically, we follow this tutorial\footnote{\url{https://docs.dgl.ai/en/0.8.x/tutorials/models/2_small_graph/3_tree-lstm.html}} to build the TreeLSTM~\citep{tai2015improved} model.

 We use Sentence-Transformer\footnote{\url{https://huggingface.co/sentence-transformers/stsb-roberta-large}}to encode the node texts into embeddings. To achieve robust validation of our methods, we run the experiments using three different random seeds and report the average and standard deviation values. The hyper-parameters used for training are provided in Table~\ref{tab:hyp_graph}. The detailed values for the ablation study experiments are provided in Table~\ref{tab:exp_ablation}.

\begin{table}[h]
    \centering
    \small
    \resizebox{5.2cm}{!}{
    \begin{tabular}{l|c}
    \toprule
        Random Seed                  & 0, 1, 2   \\
        Batch Size                   & 256   \\
        Hidden Dimension             & 128   \\
        Learning Rate   & $5\times10^{-3}$ \\
        Number of GNN Layer          & 2   \\
        Epoch                        & 20   \\
        \bottomrule
    \end{tabular}
    }
    \caption{Hyper-parameters used for experiments of customized graph methods}
    \label{tab:hyp_graph}
\end{table}

\section{Dataset Details}
\label{appx:full_example}

Here, we present the full text of 12 claims collected from a real-world patent application, each followed by its corresponding FLAN Graph.

\begin{minipage}{2.8in}
    \texttt{Claim 1:}\\
    A system for use in allowing a user to conduct one or more transactions at one or more touchpoints in a business facility, the system comprising: an authentication component configured to authenticate the user as a person allowed to conduct the one or more transactions; a tracking component configured to track the user's location within the facility as the user moves through the facility; and a control component configured to: receive authentication information from the authentication component;  receive location information from the tracking component; and  deliver a message to the touchpoint authorizing the touchpoint to engage in one or more transactions with the user. \\
    
    \includegraphics[width=\linewidth]{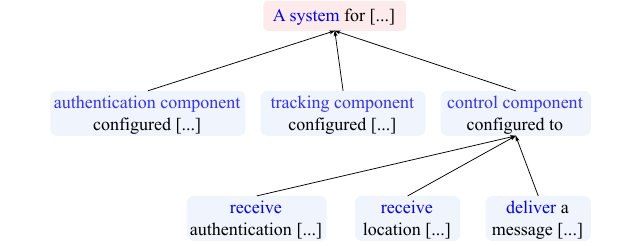}
\end{minipage}

\begin{minipage}{2.8in}
    \texttt{Claim 2:}\\
    The system of  claim 1, where the control component is also configured to use the location information to recognize that the user has moved away from the touchpoint. \\
    
    \includegraphics[width=\linewidth]{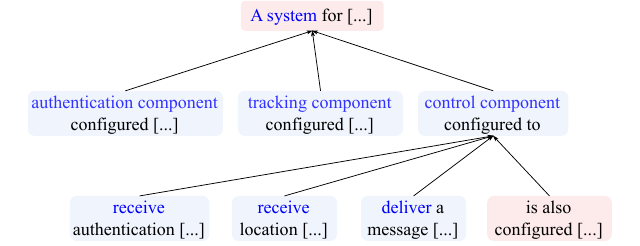}
    \end{minipage}

\begin{minipage}{2.8in}
    \texttt{Claim 3:}\\
     The system of  claim 2, where the control component is configured to deliver a second message to the touchpoint indicating that the user has moved away. \\
    
    \includegraphics[width=\linewidth]{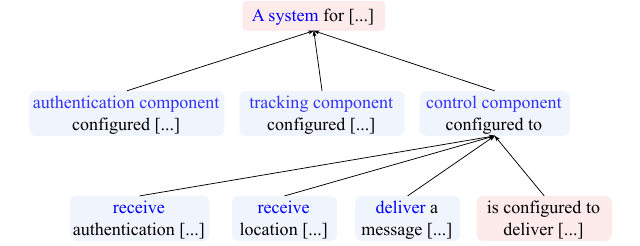}
\end{minipage}

\begin{minipage}{2.8in}
    \texttt{Claim 4:}\\
     The system of  claim 2, where the control component is configured to: use the location information to recognize that the user has moved into position to engage a second one of the touchpoints; and deliver a message to the second touchpoint authorizing the second touchpoint to engage in one or more transactions with the user. \\
    
    \includegraphics[width=\linewidth]{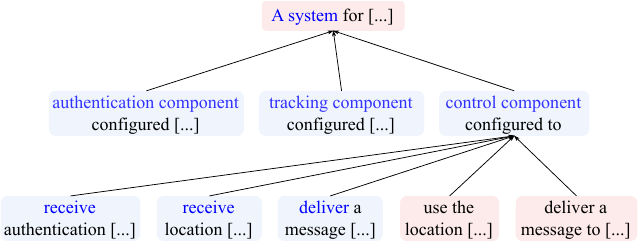}
\end{minipage}

\begin{minipage}{2.8in}
    \texttt{Claim 5:}\\
    The system of  claim 1, where the authentication component includes a terminal configured to authenticate the user when a code provided to the terminal by the user matches a code stored on a token carried by the user. \\
    
    \includegraphics[width=\linewidth]{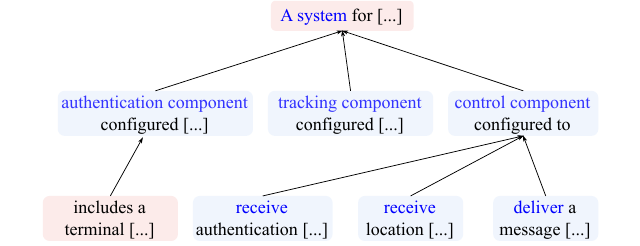}
\end{minipage}

\begin{minipage}{2.8in}
    \texttt{Claim 6:}\\
    The system of  claim 5, where the terminal is configured to receive as the token a card inserted by the user. \\
    
    \includegraphics[width=\linewidth]{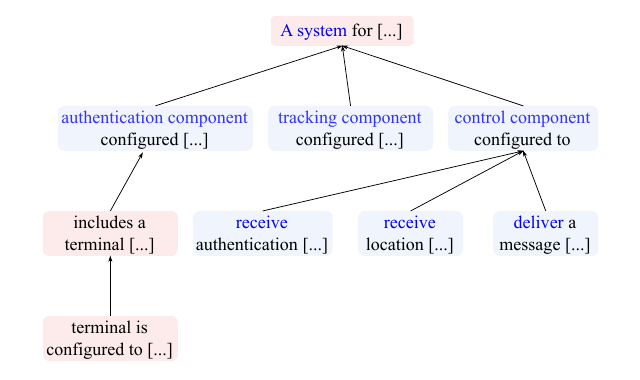}
\end{minipage}

\begin{minipage}{2.8in}
    \texttt{Claim 7:}\\
     The system of  claim 1, where the tracking component includes a visual-tracking system. \\
    
    \includegraphics[width=\linewidth]{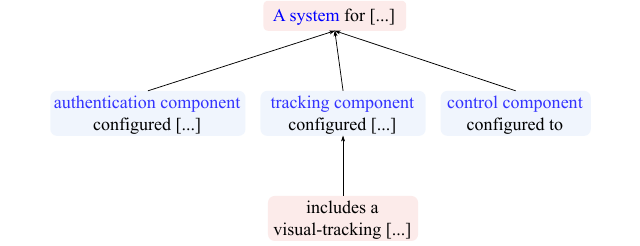}
\end{minipage}

\begin{minipage}{2.8in}
    \texttt{Claim 8:}\\
     The system of  claim 7, where the visual-tracking system includes one or more video cameras positioned within the facility.\\
    
    \includegraphics[width=\linewidth]{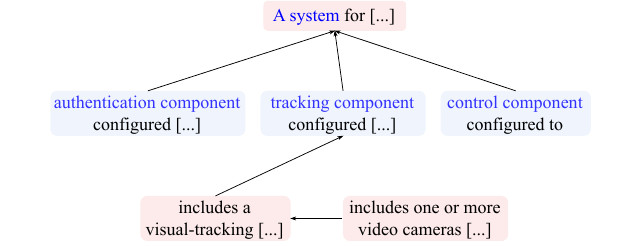}
\end{minipage}

\begin{minipage}{2.8in}
    \texttt{Claim 9:}\\
    The system of  claim 1, where the tracking component is configured to assess the users location within a grid imposed on the facility.\\
    
    \includegraphics[width=\linewidth]{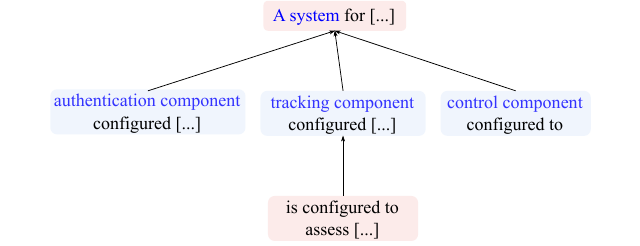}
\end{minipage}

\begin{minipage}{2.8in}
    \texttt{Claim 10:}\\
    The system of  claim 9, where the control component is configured to compare the users location within the grid to one or more fixed grid locations associated with one or more of the touchpoints.\\
    
    \includegraphics[width=\linewidth]{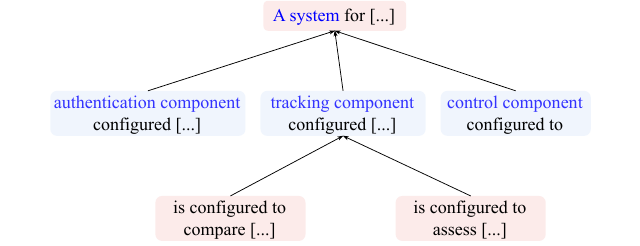}
\end{minipage}

\begin{minipage}{2.8in}
    \texttt{Claim 11:}\\
    The system of  claim 1, where the control component is configured to include information identifying the user in the message delivered to the touchpoint.\\
    
    \includegraphics[width=\linewidth]{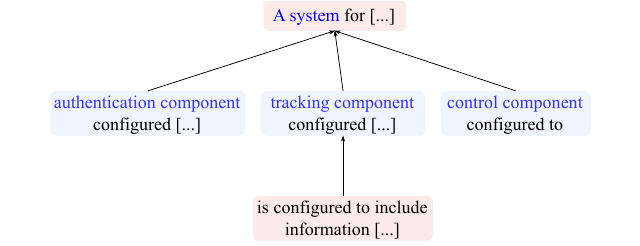}
\end{minipage}

\begin{minipage}{2.8in}
    \texttt{Claim 12:}\\
    The system of  claim 1, where the control component is configured to include an image depicting the user in the message delivered to the touchpoint.\\
    
    \includegraphics[width=\linewidth]{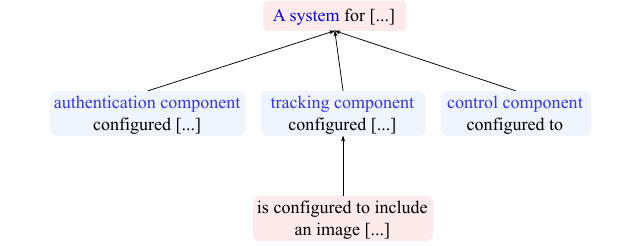}
\end{minipage}


\end{document}